\documentclass[journal]{IEEEtran}

\usepackage{fixltx2e}
\pdfoutput=1

\usepackage{amsmath}
\usepackage{multirow}
\usepackage{subfig}
\usepackage{graphicx}
\usepackage{pifont}
\usepackage{xcolor}
\newcommand{\cmark}{\textcolor{green!80!black}{\ding{51}}}
\newcommand{\xmark}{\textcolor{red}{\ding{55}}}

\usepackage{array}
\newcolumntype{L}[1]{>{\raggedright\let\newline\\\arraybackslash\hspace{0pt}}m{#1}}
\newcolumntype{C}[1]{>{\centering\let\newline\\\arraybackslash\hspace{0pt}}m{#1}}
\newcolumntype{R}[1]{>{\raggedleft\let\newline\\\arraybackslash\hspace{0pt}}m{#1}}

\usepackage{hyperref}

\usepackage{xcolor}
\definecolor{OliveGreen}{rgb}{0.16, 0.5, 0.0}
\definecolor{obblue}{rgb}{0.0, 0.47, 0.75}

\begin{document}

%
\title{
Adversarial Learning for Personalized Tag Recommendation
}

\author{Erik~Quintanilla,
        Yogesh~Rawat,
        Andrey~Sakryukin,
        Mubarak~Shah,~\IEEEmembership{Fellow,~IEEE,}
        Mohan~Kankanhalli,~\IEEEmembership{Fellow,~IEEE}
\thanks{Erik Quintanilla is with Illinois Institute of Technology, Yogesh Rawat and Mubarak Shah are with CRCV, University of Central Florida, and Andrey Sakryukin and Mohan Kankanhalli are with National University of Singapore.}}

%

\markboth{Transactions on Multimedia: \textit{under review}}%
{Equintilla \MakeLowercase{\textit{et al.}}: Adversarial Learning for Personalized Tag Recommendation}

\maketitle

%
\begin{abstract}

We have recently seen great progress in image classification due to the success of deep convolutional neural networks and the availability of large-scale datasets. Most of the existing work focuses on single-label image classification. However, there are usually multiple tags associated with an image. The existing works on multi-label classification are mainly based on lab curated labels. Humans assign tags to their images differently, which is mainly based on their interests and personal tagging behavior. In this paper, we address the problem of \textit{personalized} tag recommendation and propose an end-to-end deep network which can be trained on large-scale datasets. The user-preference is learned within the network in an \textit{unsupervised} way where the network performs joint optimization for user-preference and visual encoding. A joint training of user-preference and visual encoding allows the network to efficiently integrate the visual preference with tagging behavior for a better user recommendation. In addition, we propose the use of adversarial learning, which enforces the network to predict tags resembling user-generated tags. We demonstrate the effectiveness of the proposed model on two different large-scale and publicly available datasets, YFCC100M and NUS-WIDE. The proposed method achieves significantly better performance on both the datasets when compared to the baselines and other state-of-the-art methods. The code is publicly available at \href{https://github.com/vyzuer/ALTReco}{https://github.com/vyzuer/ALTReco}.
\end{abstract}


%
%

%

%

%

\IEEEpeerreviewmaketitle

\section{Introduction}
We have seen a large number of emerging social media platforms where users share their experiences with others leading to big multimedia data \cite{tian2015multimedia, zhu2015multimedia}. Millions of photographs are daily shared by social media users on these platforms. The presence of tags, along with these photographs, allow other users to search for these shared images easily and is also important for various recommendation services \cite{xu2014graph, sun2016social, cui2016social, wu2017inferring, yao2017joint}. Assigning tags to these photographs before sharing on these social media platforms can be challenging as well as time-consuming for most users. Therefore, tag recommendation is an interesting research problem for the multimedia research community. In this work, we focus on automatic tag recommendation, which will assist users in tagging their photographs.

\begin{figure}[t!]
\centering
\captionsetup{justification=centering}
\centering\includegraphics[width=0.8\linewidth]{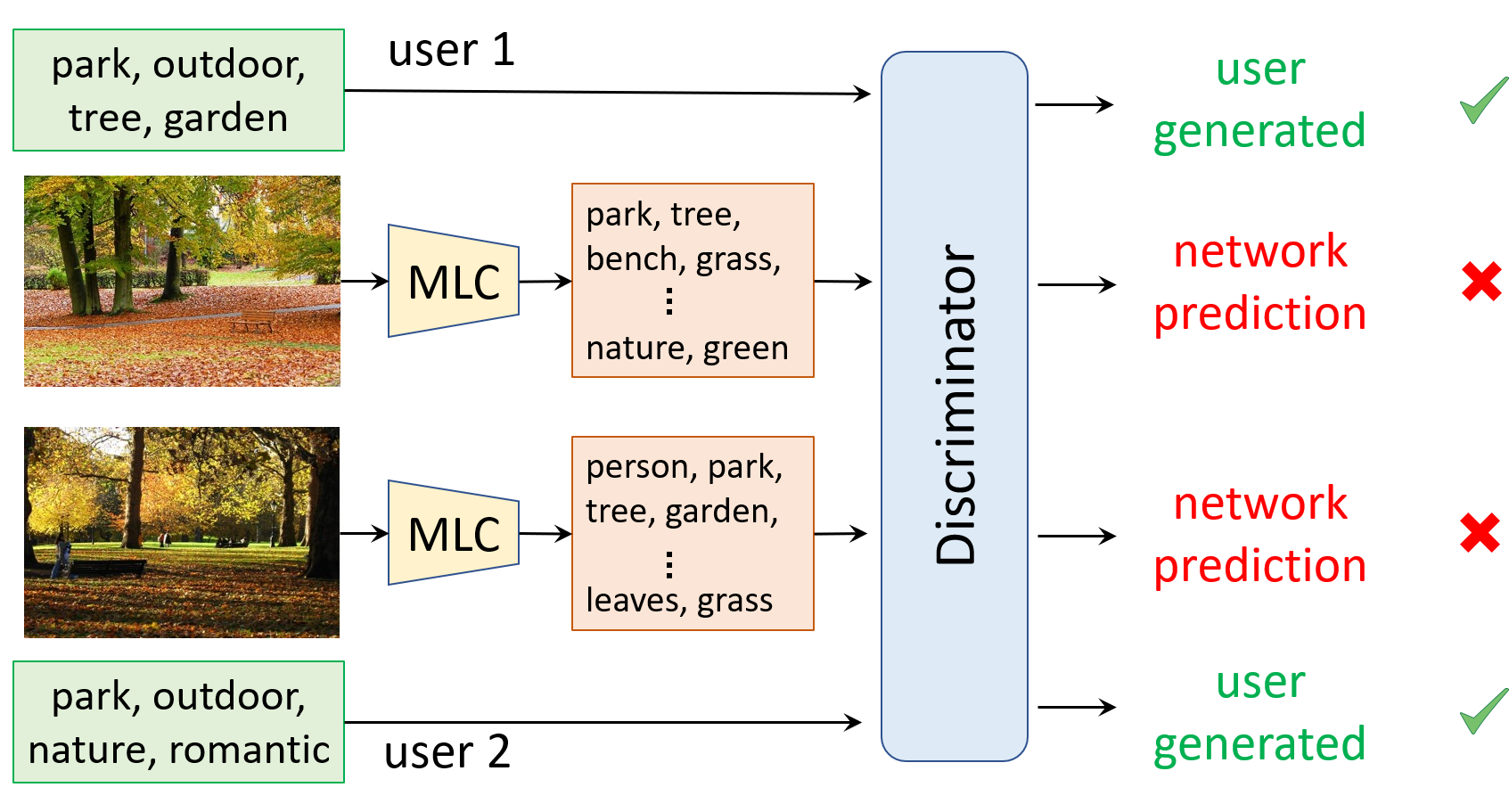}
\caption{\small{Humans annotate their images differently as compared to a multi-label image classifier (MLC), which lists all the visual objects present in the image. Here we show two images from YFCC100M \cite{yfcc100m} 
along with user-assigned tags, which are diverse as well as different. We can also see the list of tags predicted using a classifier. We propose to use a discriminator as an adversary, which specializes in differentiating between user-generated and network predicted tags. This enables the network to predict preference-aware tags resembling user-generated tags.}}
\label{fig_overview}
\end{figure}

In recent years, tremendous improvement has been reported in visual classification due to the success in deep convolutional neural networks (CNNs) \cite{krizhevsky_2012_imagenet, simonyan_2014_very, szegedy_2015_going, liao2015tag, cui2017general, fang2018attention, xie2019convolutional, li2019coco}. A single label is often not sufficient to describe the whole image, and therefore multiple tags are usually assigned by the users to their images. This has been explored via a multi-label image classification problem, where an image can have multiple user-tags or labels \cite{Makadia_2008_NBI, Guillaumin_2009_tagprop, Weston_2010_large_scale, li_2014_multi, huang2015unconstrained, ding2016multi, wang_2016_cnn, liu2018multi}. However, these works are mainly focused on datasets that were manually annotated based on the visual content, such as ImageNet \cite{deng_2009_imagenet} and NUS-WIDE \cite{chua_2009_nus} (81 visual concepts). Also, the different sets of visual concepts are determined by researchers in a lab environment and not by the real users, which makes them less relevant for image tag recommendation. A classifier trained to identify visual concepts in an image will merely list all the objects present in the image (Figure \ref{fig_overview}). Humans, on the other hand, annotate their images differently. Therefore, the tags predicted by a classifier in this manner may not be well suited for user recommendation.

We propose the use of adversarial learning to address this issue. The goal is to train the recommendation network to learn effective features which can predict a list of tags which resembles human-generated tags for any given image. A discriminator, which specializes in differentiating between machine-generated and human-generated tags, is used as an adversary for the recommendation network. The recommendation network is penalized if the discriminator is able to distinguish between the predicted and human-generated tags. The network is trained by optimizing this loss jointly with multiple other losses, and eventually, it learns to predict the tags which resemble human-generated tags.

Personalization also plays a significant role in tagging, as different users may have different preferences and interests \cite{liu2014personalized, zhang2017personalized, zhang2019personalized}. We can have an entirely different set of tags present in similar looking images from different users (Figure \ref{fig_overview}). Therefore, it is important for a system to consider both visual content as well as user preference when recommending tags to a user. This is not only useful to the individual user, but it also provides diversity and indirectly helps in search and recommendation services. Most of the existing methods for personalized tag recommendation make use of tag co-occurrence \cite{sigurbjornsson2008flickr}, matrix factorization \cite{cai2011low, rafailidis2013tfc, rendle2010pairwise}, and graph-based models \cite{guan2009personalized}. However, these methods work well on smaller datasets and do not scale well with large-scale datasets. In this work, we propose a deep network which integrates user-preference with the visual content of the image and is well suited for large-scale datasets. 


We make the following contributions in this work:
\begin{enumerate}
    \item We propose an end-to-end deep network for preference-aware tag recommendation, which can be trained on large-scale datasets. We use an encoder-decoder network equipped with skip connections which enables efficient and \textit{unsupervised} user-preference learning. A joint training of user-preference and visual encoding allows the network to efficiently integrate the visual preference with tagging behavior for a better recommendation.
    \item We propose the use of \textit{adversarial} learning to further enhance the quality of tag prediction. The adversarial loss enforces the network to learn features which can be used to predict tags with a distribution which resembles human-generated tags.
    \item We perform extensive experiments on two different datasets, YFCC100M and NUS-WIDE. The proposed method achieve significant improvement on the large-scale YFCC100M dataset and also outperforms existing methods on NUS-WIDE dataset.
\end{enumerate}


\begin{figure}[t!]
\centering
\captionsetup{justification=centering}
\begin{subfloat}[][\textbf{sunset}, \textbf{clouds}, \textbf{sky}]{
  \centering\includegraphics[width=0.2\linewidth, height=0.15\linewidth]{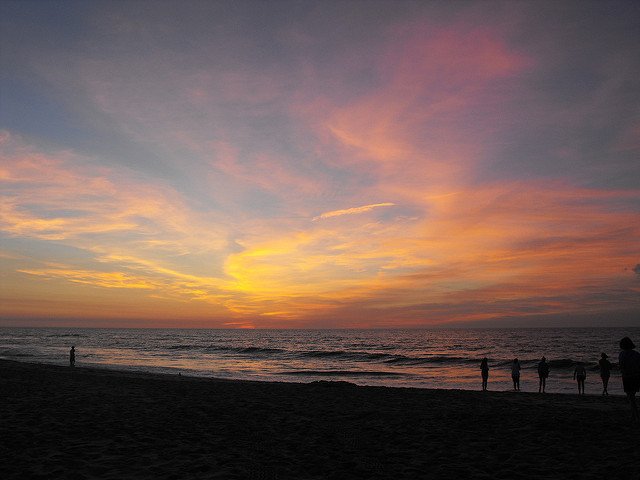}}
\end{subfloat}
\begin{subfloat}[][\textbf{ocean},\textbf{beach}, \\ \textbf{shore},\textbf{coast}]{
  \centering\includegraphics[width=0.2\linewidth, height=0.15\linewidth]{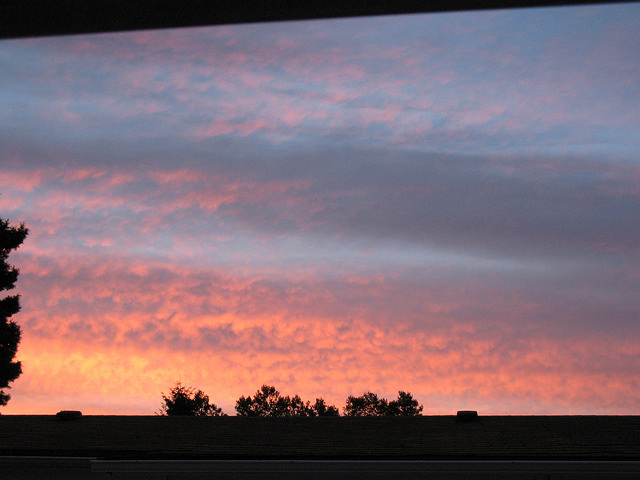}}
\end{subfloat}
\begin{subfloat}[][aerial, alps, mountain, panoramic]{
  \centering\includegraphics[width=0.2\linewidth,height=0.15\linewidth]{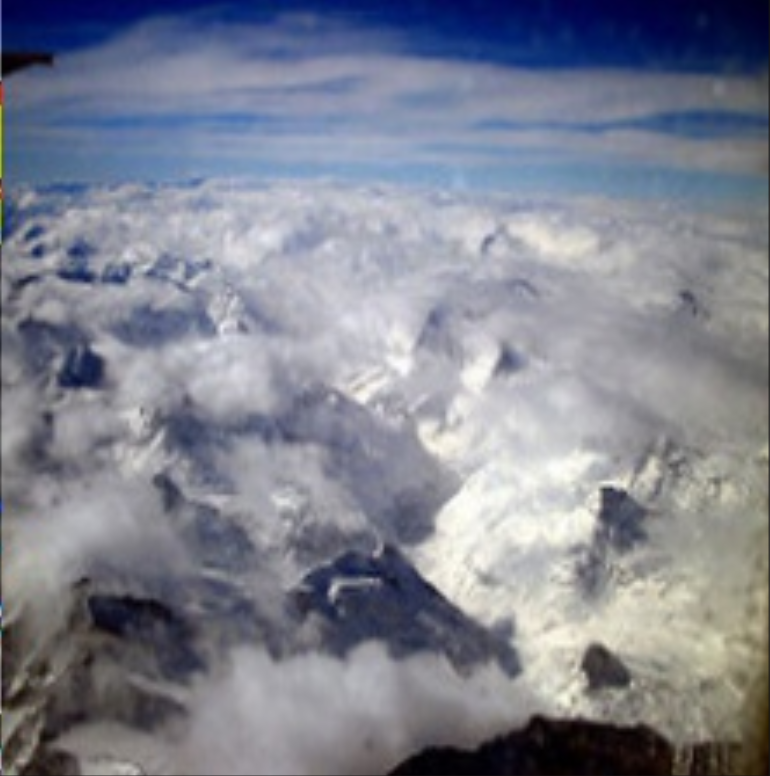}}
\end{subfloat}
\begin{subfloat}[][above, aerial, alps, landscape, snow, view]{
  \centering\includegraphics[width=0.2\linewidth, height=0.15\linewidth]{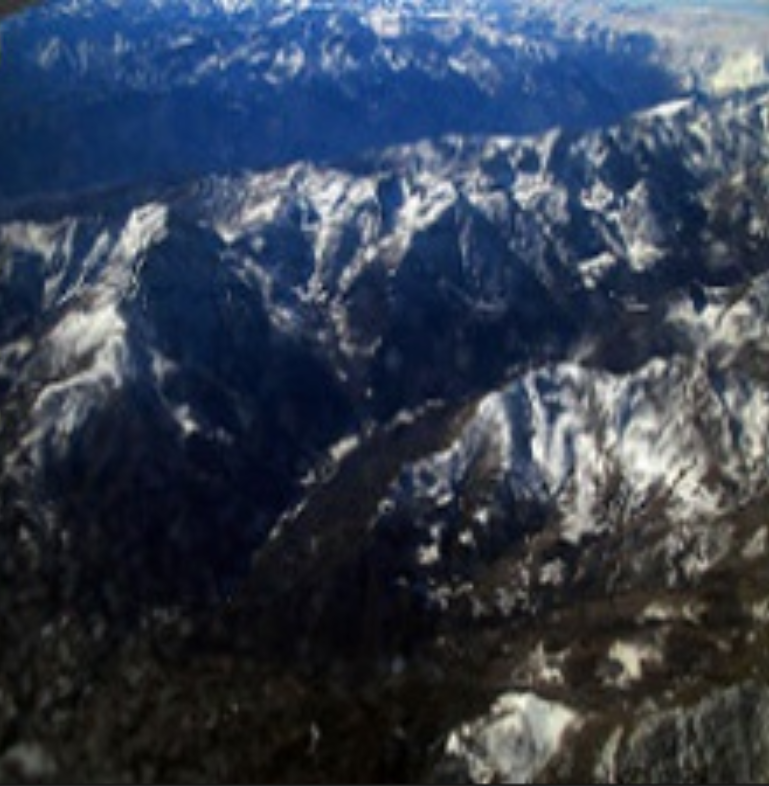}}
\end{subfloat}
\caption{\small{Some sample images from YFCC100M dataset \cite{yfcc100m} along with user assigned tags. Each image has been assigned  multiple user tags. It can be observed that the first two images (a,b) and the last two images (c,d) have almost similar visual content, but the owners have assigned different set of tags to these images.}}
\label{fig_sample_pics}
\end{figure}

\section{Related Work}
\label{lb_related}
Our proposed approach mainly overlaps with three research areas: multi-label classification, personalized recommendations, and adversarial learning. 

\textbf{Multi-label classification.} 
In recent years, we have seen great progress in image classification with the advancement in deep learning and the introduction of Convolutional Neural Networks (CNNs) \cite{krizhevsky_2012_imagenet}. 
Important to mention, this progress was also made possible due to the release of large-scale annotated datasets such as ImageNet \cite{deng_2009_imagenet}. 
The availability of multi-label dataset, such as VOC PASCAL \cite{everingham2010pascal}, NUS-WIDE \cite{chua_2009_nus}, and Microsoft COCO \cite{lin2014microsoft} has also led to further development of multi-label classification research.

Early deep learning approaches focused mostly on the design of better loss functions and network architectures.
So, Gong et al. \cite{gong2013deep} investigated multiple ranking based cost functions and proposed a weighted approximated ranking loss to solve multi-label image annotation. 
Weston et al. \cite{weston2011wsabie} proposed to optimize the top-of-the-list ranking to learn a common embedding for images and labels. 
On a similar path, the authors in \cite{frome2013devise} proposed a unified model for visual and semantic joint embedding which utilizes image semantics along with pixel values to identify visual objects. 
Wei et al. \cite{wei2014cnn} proposed the use of max-pooling to aggregate the prediction from multiple hypothesis in Hypothesis-CNN-Pooling (HCP). 
In \cite{gong2014multi}, the authors proposed canonical-correlation-analysis (CCA) to learn a common latent space for joint mapping of visual features and labels. 

The recent works, in contrast, focus more on producing better quality results by taking into account inter-class correlations. 
The authors in \cite{wang_2016_cnn} proposed a deep network combining a recurrent and a convolutional network.
In this setting, the model learns a joint image-tag embedding, which takes into account inter-class dependency along with visual features and classes relations. 
Similarly,  in \cite{li2018attentive}  Attention layer, along with RNN, is applied to ensure that generated classes refer to different aspects of the image. 
Authors of \cite{chen2017graph} apply Graph Convolutional Networks for learning of the relations between different tags in the form of a graph structure.
Alternatively,  uses of Graph Gated NNs with external knowledge as an approach to zero-shot learning in the multi-label classification task is applied in \cite{lee2018multi}. 

\begin{figure*}[t!]
\centering
\captionsetup{justification=centering}
\centering\includegraphics[width=0.85\linewidth]{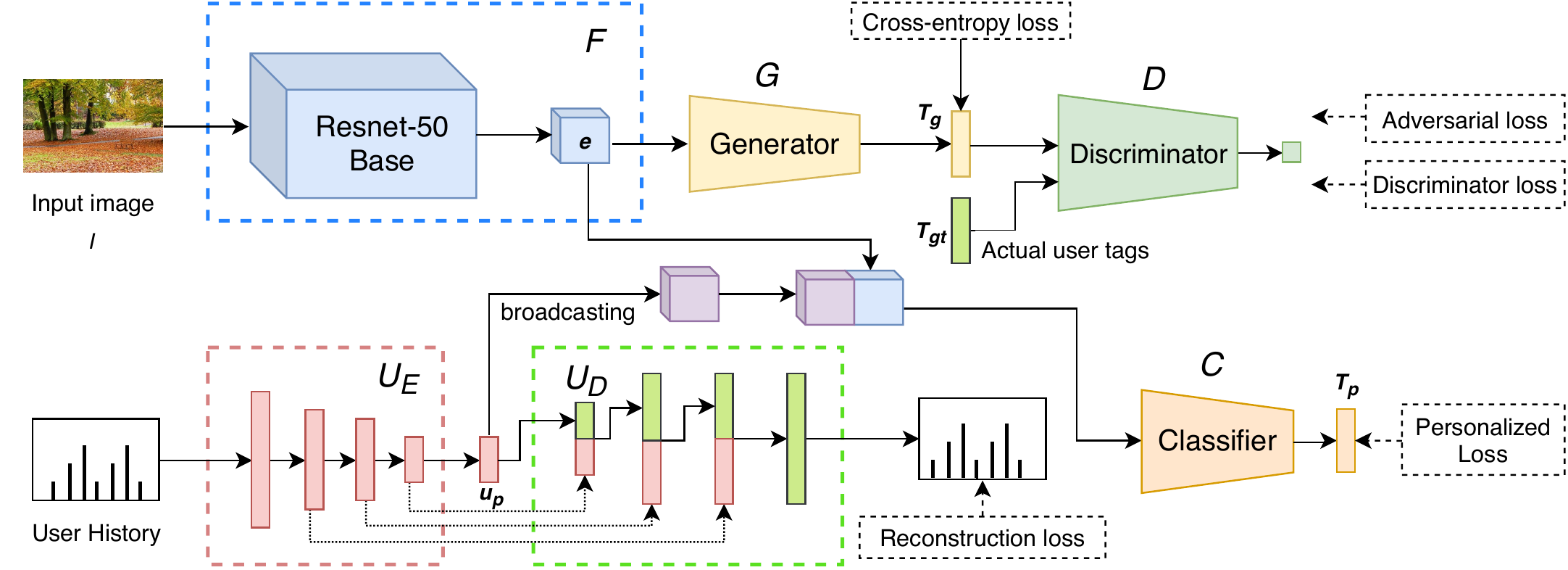}
\caption{\small{Overview of the proposed framework. Input image \textit{I} is encoded using a convolutional neural network \textit{F} and integrated with the user embedding learned via unsupervised auto-encoder \textit{U\textsubscript{E}} and \textit{U\textsubscript{D}}. The preference conditioned embedding is used for personalized tag prediction using a classifier \textit{C}. The generator \textit{G} takes the visual embedding and predict tags which are passed to a discriminator \textit{D} for adversarial learning.}}
\label{fig_framework}
\end{figure*}

\textbf{Personalized recommendations.} 
Most of the previous research on personalized recommendations were using more classical approaches not involving deep learning.
One of the popular approaches is the creation of a user-item interaction matrix and its further factorization.
For instance, authors of \cite{cai2011low} proposed a novel algorithm using Low-Order Tensor decomposition, while \cite{rendle2010pairwise} aimed to achieve better performance and scalability of existing approach through the introduction of Tensor Factorization technique. 
Further, \cite{rafailidis2013tfc} achieved better performance by using High Order Singular Value Decomposition Tensor Factorization technique with a tag clustering stage. 
Rae et al.~\cite{rae2010improving} in their work use probabilistic model utilizing user tagging preference as well as his social contacts and social groups. 
Similarly, \cite{shah2016prompt} calculates user tagging behavior (UTB) vectors and uses them along with UTB vectors of neighboring users as an input to the model.
Individual user tagging history has been studied before with application to Factorization Machines in \cite{nguyen2017personalized}.
Alternatively, \cite{mcparlane2013contextual} used the tagging history data corresponding to the image location and time. 
Further, the role of context has been explored in \cite{su2013flickr} and \cite{johnson2015love}, these works are mainly focused on single tag prediction\cite{su2013flickr} and are trained on smaller datasets with a small set of lab curated labels \cite{johnson2015love}. 
In the deep learning setting, \cite{nguyen2017personalized2} created a dictionary of users, which were fed to the network using one-hot encoding. 

In contrast to the existing works, here we propose a novel deep learning multi-tagging model utilizing user-specific historical data.
We investigate the role of user-preference jointly with visual content for tag recommendation.
Unlike other approaches, our work utilizes real-world data with a large number of users and images.

\textbf{Adversarial learning.} 
Application of adversarial learning strategies showed tremendous success in the Image Generation research \cite{ledig2017photo,zhang2017stackgan,liu2016coupled, fang2019learning}.
However, their application in multi-label tagging is a relatively unexplored task.
Adversarial learning was previously applied in the text domain by \cite{babbar2018adversarial}.
In the image domain, authors in \cite{wu2018tagging} treated the image classification part as a generator, while discriminator was used to classify produced tags as relevant or non-relevant for the input image. In \cite{mirza2014conditional}  conditional GAN network for multi-label classification was used, where the semantic embedding of tags was used for discrimination. In contrast to these, in our work, we use the co-occurrence of predicted tags for adversarial learning which in turn enables the network to learn better features for predicting tags which resemble human behavior.

\section{Personalization}
There are two different aspects of personalized image tagging. The first one corresponds to the choice of labels, which we refer to as {\em tagging behavior}. And, the other corresponds to a preference for visual content which we term as {\em visual-preference}. Different users may have different image-tagging behavior, and we propose to model this aspect based on the past tagging behavior of the users. Existing works which focus on tagging behavior for personalized recommendation utilize collaborative filtering \cite{cai2011low, rafailidis2013tfc, rendle2010pairwise} and graph-based approaches \cite{guan2009personalized}. These approaches are either not scalable for large-scale datasets or they ignore the visual content for the recommendation and are mostly based on tag co-occurrence. We propose a deep auto-encoder based approach which is well suited for large-scale datasets and the user-preference is learned in an unsupervised way.

We observe that different users assign different tags to visually similar images. Figure \ref{fig_sample_pics} shows some example images from YFCC100M dataset from different users. The first two examples are for sunset images and are visually very similar. We can observe that the first user is interested in tags such as \textit{sunset}, \textit{clouds} and \textit{sky}, whereas, the other user has tagged the image as \textit{ocean}, \textit{beach}, and \textit{coast}. Similarly, in example (c) and (d) we can observe that the first user is interested in tags such as \textit{mountain} and \textit{panoramic}, whereas the second user has used tags \textit{landscape} and \textit{snow} to describe similar looking image. This shows that the visual content of an image is not sufficient for a personalized tag recommendation.

Apart from varying tagging behavior, this also reflects the fact that different users may focus on different visual aspects of the image. The user for image (a) is interested in \textit{clouds} and \textit{sky}, whereas the user for image (b) is more interested in \textit{beach} and \textit{coast}. This variance in tagging behavior also poses a challenge for tag prediction in real-world data without considering user-preferences. We address this aspect of personalization by performing joint learning of user-preference along with visual encoding of the input image. The key idea is to learn to focus on those visual aspects of the input image, which are personalized to a specific user. The joint optimization of the proposed network for user-preference and visual encoding at the same time allows the network to learn this aspect of personalization.

\section{Proposed Method}
\label{lb_method}
We propose a preference-aware deep network which solves a multi-label image classification problem. A detailed overview of the proposed network is shown in Figure \ref{fig_framework}. The proposed network takes an image $I$ as input along with a user tagging history $\boldsymbol{u}_h$ and predicts a list of personalized tags $\boldsymbol{T}_p$. The input image $I$ is encoded using a convolutional neural network $\mathcal{F}(I)$ to get a visual descriptor $\boldsymbol{e}$. The visual descriptor $\boldsymbol{e}$ is integrated with a latent user-preference $\boldsymbol{u}_p$,  and a classifier $\mathcal{C}(\boldsymbol{e}, \boldsymbol{u}_p)$ take these encodings to predict personalized tags $\boldsymbol{T}_p$. The user-preference $\boldsymbol{u}_p$ is learned in an unsupervised way using an auto-encoder, which consists of an encoder $\mathcal{U_E}$ and a decoder $\mathcal{U_D}$. The network also has a generator $\mathcal{G}$, which takes the visual encoding $\boldsymbol{e}$ and predicts generalized tags $\boldsymbol{T}_g$ with adversarial learning using a discriminator $\mathcal{D}$. The discriminator $\mathcal{D}$ acts as an adversary to $\mathcal{G}$ and learns to distinguish between the generated and ground truth image tags.

The network is composed of two parallel input streams, one for visual encoding $\mathcal{F}(I)$ and the other for integration of user-preference $\mathcal{U_E}(\boldsymbol{u}_h)$. We have adopted ResNet-50 \cite{he2016deep} for encoding the visual information in the input image. We take the activation feature maps from layer 46 (\textit{activation\_46}) of the last ResNet block as visual encoding $\boldsymbol{e} \in \mathcal{R}^{h\times w \times c}$. It is important to note that the visual encoding network $\mathcal{F}(I)$ can be easily substituted with any other state-of-the-art convolutional networks such as InceptionNet \cite{szegedy2015going}, DenseNet \cite{huang2017densely}, etc. The visual encoding $\boldsymbol{e}$ is used to predict personalized image tags ($\boldsymbol{T}_p$) and also generate generalized tags ($\boldsymbol{T}_g$) for adversarial learning.

\paragraph{\textbf{User-preference}}
\label{lb_upref}
The user-preference $\boldsymbol{u}_p$ is learned using an auto-encoder where the encoder $\mathcal{U_E}(\boldsymbol{u}_h)$ learns a latent representation $\boldsymbol{u}_p$ and the decoder $\mathcal{U_D}$ reconstructs the user tagging history $\boldsymbol{u}_h$. We utilize the user tags associated with each user in their past captured images to define user tagging history $\boldsymbol{u}_h$. A user tagging history is defined as $\boldsymbol{u}_h = [u^0, u^1, ... , u^N]$, where $u^i$ indicates the presence of tag $i$ in the user's history of tags normalized by the maximum count of any tag present in user history. Here $N$ represents the size of the tag vocabulary in the dataset. It is important to note that the user history is different from the class labels used for classifier as it accounts for all the images from a particular user. One the other hand, class labels used for training the classifier are specific to the image used as input. 

The proposed network for learning user-preference consists of two components, an encoder $\mathcal{U_E}$ and a decoder $\mathcal{U_D}$. The encoder $\mathcal{U_E}$ takes user history vector $\boldsymbol{u}_h$ as input and generates a latent representation $\boldsymbol{u}_p$ as user-preference. It consists of a series of 4 fully connected layers with $1024, 512, 256$, and $128$ neurons each with $relu$ activation. The latent representation $\boldsymbol{u}_p$ is a 128-dimensional vector indicating the preference of a user. The decoder $\mathcal{U_D}$ takes this representation $\boldsymbol{u}_p$ and reconstructs the user history vector $\boldsymbol{u}_h$. Reconstruction of the user history vector $\boldsymbol{u}_h$ can be challenging with this approach. We want the encoder network $\mathcal{U_E}$ to learn similar latent representations for users with similar tagging behavior. If the encoder network $\mathcal{U_E}$ generates similar latent representations for two different users, the decoder network $\mathcal{U_D}$ will not be able to reconstruct the exact user history vectors $\boldsymbol{e_h}$ using the same latent representation $\boldsymbol{u_p}$. 

We address this issue in two steps. In the first step, we take inspiration from U-net architecture \cite{ronneberger2015u} and propagate user history encodings from the encoder network $\mathcal{U_E}$  to the decoder network $\mathcal{U_D}$ via skip connections. The encodings from the encoder are concatenated with the embeddings in the corresponding decoder layer. This ensures the availability of user-specific encodings in the decoder network, which facilitates the reconstruction of user history vector $\boldsymbol{u_h}$. We want to learn similar latent representation for users with similar tagging behavior. However, even for users with similar tagging behavior, there will be a lot of variations in tag usage, such as in terms of frequency. Therefore, two different users, even with similar tagging behavior, will not have identical tagging history. The user-preference is learned using user-history reconstruction in an unsupervised way. If we don’t have skip connections, it will be hard for the network to learn similar latent representation for users with similar tagging behavior as the latent representation should have sufficient information to perform the reconstruction, which might be different for different users even if they have similar tagging behavior. The skip connections allow user-specific details (user variations) to pass from encoder to decoder, which will help in the reconstruction. Therefore the skip connections allow the encoder network $\mathcal{U_E}$ to generate a meaningful latent representation for user-preference and jointly train $\mathcal{U_E}$ and $\mathcal{U_D}$ in an unsupervised way. The decoder network $\mathcal{U_D}$ consists of a sequence of 4 fully connected layers with 128, 256, 512, and 1024 neurons, each with $relu$ activation. Each layer also takes encodings from the corresponding layer in the encoder network. The last layer is followed by one more fully connected layer, which reconstructs the user history vector $\boldsymbol{\hat{u}_h}$ using a sigmoid activation.

In the second step, we ensure that the reconstruction of the user history vector $\boldsymbol{u_h}$ can be easily approximated by $\mathcal{U_D}$. There will be a large variation in the user history vector $\boldsymbol{u_h}$ even for the users with similar latent representation $\boldsymbol{u_p}$. Therefore, we propose the use of Huber loss \cite{huber1992robust} which is less sensitive to outliers. We optimize the auto-encoder with the following loss function,
\begin{align}
L_r = -\frac{1}{N}\sum\limits_{i=1}^{N}\left\{ \begin{array}{cl}
\frac{1}{2} \left[u^i-\hat{u}^i\right]^2 & \text{for }|u^i-\hat{u}^i| \le \delta, \\
\delta \left(|u^i-\hat{u}^i|-\delta/2\right) & \text{otherwise.}
\end{array}\right.
\end{align}
where $L_r$ is the reconstruction loss and $N$ is the size of the tag vocabulary, $\hat{u}^i$ is the ground-truth tag presence indicator in the user history and ${u}^i$ is the corresponding prediction. $\delta$ is a constant threshold which we set to 1.0 in all our experiments.

\paragraph{\textbf{Personalized Tags}}
\label{lb_pc_cnn}
The visual encoding $\boldsymbol{e}$ from the encoder network $\mathcal{F}$ is integrated with the user-preference $\boldsymbol{u_p}$ learned using the preference encoder network $\mathcal{U_E}$. The integrated encodings are then passed to a multi-label classifier $\mathcal{C}$ which predicts personalized tags. The classifier network $\mathcal{C}$ consists of two convolution layers followed by two fully connected layers. We optimize the personalized tag prediction with a cross-entropy loss,
\begin{equation}
L_p = -\frac{1}{N}\sum\limits_{i=1}^{N}\hat{p}_{i}log({p}_{i}) + (1-\hat{p}_{i})log(1-{p}_{i}),
\end{equation}
where $\hat{p}_{i}$ indicates the presence or absence of a tag in the image ground truth and ${p}_{i}$ is the network prediction $\mathcal{C}(\boldsymbol{e}, \boldsymbol{u_p})$ corresponding to that tag. $L_p$ represents the loss corresponding to personalized tags and $N$ denotes the size of the tag vocabulary in the dataset.

We experimented with different integration techniques, including addition, concatenation, and multiplication. We observe that the concatenation operation outperforms the other two variations. This is intuitive as concatenation allows the network to learn how to integrate these two encodings for a better prediction. The concatenation operation also enables the network to use the user-preference independently. This is important as tagging behavior will also play an important role in tag recommendation along with the visual content of the image.

\subsection{Adversarial Learning}
Adversarial learning has recently been applied to a wide range of problems \cite{goodfellow2014generative, donahue2016adversarial}. It has shown a tremendous success mainly in image generation tasks \cite{zhu2017unpaired, karras2017progressive}. In this work, we propose to explore the adversarial learning for tag recommendation. The idea is to use an additional adversarial loss for tag prediction, which ensures that the predicted tags have a distribution which resembles actual user-generated tags. A classifier trained on images will usually predict all the tags corresponding to the visual content of the image. This is, however, not useful for a recommendation, as it will merely list all the visual objects present in the image. Humans, on the other hand, describe an image differently with their own set of tag vocabulary and personal interests. The introduced adversarial loss penalizes the network prediction if the predicted tags are not from a distribution similar to user-generated tags.

We incorporate the adversarial loss in a Generative Adversarial Network (GAN) \cite{goodfellow2014generative} framework, where the tag predictions from generator network, $\mathcal{G}$, are considered as generated labels and there is a discriminator, $\mathcal{D}$, which differentiates between generated and ground truth user-generated tags. The goal of the generator network, $\mathcal{G}$, is to fool the discriminator network, $\mathcal{D}$, and generate tags which resemble user-generated tags. This is achieved by minimizing the following objective function,
\begin{equation}
\begin{aligned}
L_{adv} &= E_{e \sim \mathcal{F}(I)}[log(1 - \mathcal{D}(\mathcal{G}(e)))]
\end{aligned}
\label{equ:G_objective}
\end{equation}
where $L_{adv}$ represents the adversarial loss and $\mathcal{F}$ is the visual encoder. The generator network $\mathcal{G}$ takes the visual encoding $\boldsymbol{e}$ as input and predicts generalized tags $\boldsymbol{T}_g$. The generator network $\mathcal{G}$ consists of three convolution layers followed by one fully connected layer for tag prediction. 

It is important to note that the adversarial loss is applied on the predicted generalized tags $\boldsymbol{T}_g$ and not on the personalized tags $\boldsymbol{T}_p$. The generator $\mathcal{G}$ does not have any interaction with the user-preference $\boldsymbol{u_p}$, and therefore it predicts generalized tags with no notion of personalization. The integration of user-preference, $\boldsymbol{u_p}$, to predict personalized tags using the classifier network $\mathcal{C}$ bridges the gap between ground truth user-generated tags and the predicted personalized tags. This makes it hard for the discriminator $\mathcal{D}$ to differentiate between user-generated and predicted tags. Therefore the adversarial loss is not found to be very effective when computed over the personalized tags. Also, the network is trained jointly in an end-to-end fashion for both tag prediction and preference learning. If the adversarial loss is computed on personalized tags, it also affects the learning of user-preference. We found this unfavorable for both tag prediction as well as the unsupervised learning of user-preference. Therefore, adversarial learning is performed on a separate branch with a generator network $\mathcal{G}$, which does not interact with user-preference.

In addition to the adversarial loss, the generator network $\mathcal{G}$ also has a cross-entropy objective function defined as,
\begin{equation}
L_g = -\frac{1}{N}\sum\limits_{i=1}^{N}\hat{p}_{i}log({p}_{i}) + (1-\hat{p}_{i})log(1-{p}_{i}),
\end{equation}
where $\hat{p}_{i}$ indicates the presence or absence of a tag in the image ground truth and ${p}_{i}$ is the network prediction $\mathcal{G}(\boldsymbol{e})$ corresponding to that tag. $L_g$ represents the loss corresponding to generalized tags and $N$ denotes the size of the tag vocabulary in the dataset. The full network consists of a visual encoder $\mathcal{F}$, user-preference encoder $\mathcal{U_E}$, user-preference decoder $\mathcal{U_D}$, a multi-label classifier $\mathcal{C}$, and a generator $\mathcal{G}$, and is trained end-to-end using the following losses,
\begin{equation}
L_t = \alpha L_p + \beta L_g + \gamma L_r + \theta L_{adv},
\end{equation}
where $L_t$ is the total network loss, and $\alpha, \beta, \gamma$ and $\theta$ are weights for these losses. We use equal weights for all the losses in our experiments.

\begin{table*}[t!]
\centering
\scriptsize
\setlength\tabcolsep{1pt} 
\begin{tabular}{ | C{2.0cm} |  C{2.0cm} | C{2.0cm} | C{2.0cm} | C{2.0cm} | C{2.0cm} |C{2.0cm} |C{2.0cm} |}
\multicolumn{1}{c}{\includegraphics[height=0.08\linewidth, width=0.11\linewidth]{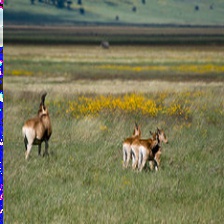}} &
\multicolumn{1}{c}{\includegraphics[height=0.08\linewidth, width=0.11\linewidth]{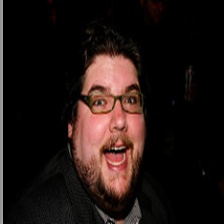}} & \multicolumn{1}{c}{\includegraphics[height=0.08\linewidth, width=0.11\linewidth]{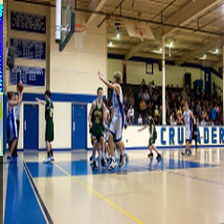}} &
\multicolumn{1}{c}{\includegraphics[height=0.08\linewidth, width=0.11\linewidth]{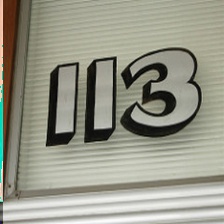}} & \multicolumn{1}{c}{\includegraphics[height=0.08\linewidth, width=0.11\linewidth]{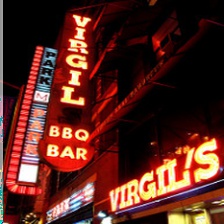}} &
\multicolumn{1}{c}{\includegraphics[height=0.08\linewidth, width=0.11\linewidth]{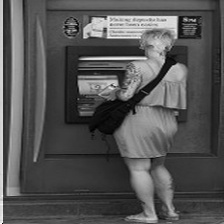}} & 
\multicolumn{1}{c}{\includegraphics[height=0.08\linewidth, width=0.11\linewidth]{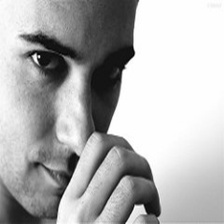}} &
\multicolumn{1}{c}{\includegraphics[height=0.08\linewidth, width=0.11\linewidth]{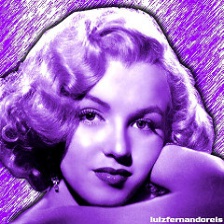}} \\ \hline
{\color{black} crater, fun, group, safari, travel, trip, view, wildlife} & 

{\color{black} beard, man, people, portrait, smile}  &
{\color{black} action, alpha, basket, basketball, sport, team}  & 
{\color{black} color, sign, signage, type, typography}  & 
{\color{black} awesome, bar, city, downtown, interesting, neon, night}  & 
{\color{black} black, downtown, legs, people, street+photography} & 
{\color{black} eyes, face, grain, hand, portrait, smile} & 
{\color{black} art, beautiful, blonde, cinema, color, glamour, peace, pop, studio } \\  \hline

{\color{OliveGreen} travel, wildlife, safari, group, trip}&

{\color{OliveGreen} portrait, man, people, smile, } {\color{obblue} face }  &
{\color{OliveGreen} alpha, sport, team, basketball, action }  & 
{\color{OliveGreen}  color, sign, } {\color{red} travel, }{\color{OliveGreen} typography, }{\color{obblue} digital}& 

{\color{obblue} neon+sign }{\color{OliveGreen} , interesting, neon, night, } \newline {\color{red} cinema} &

{\color{obblue} black+and+white, }{\color{OliveGreen} street+photography, } {\color{red} street, man, }{\color{OliveGreen}downtown}  &
{\color{OliveGreen}portrait, face, eyes, }{\color{obblue}boy, male}& 
{\color{OliveGreen}glamour, studio, }{\color{obblue}portrait, vintage, face} 
\\ \hline
\end{tabular}
\caption{\small{Some example images from YFCC100M dataset with the ground-truth tags in the first row and the top-5 recommended tags in the second row. The tags in green match the ground truth, tags in blue are relevant but not present in the ground truth, and the red ones are wrong predictions.}}
\label{table_sample_results_yfcc1}
\end{table*}

\begin{table*}[t!]
\centering
\scriptsize
\setlength\tabcolsep{1pt} 
\begin{tabular}{ | C{2.0cm} |  C{2.0cm} | C{2.0cm} | C{2.0cm} | C{2.0cm} | C{2.0cm} | C{2.0cm} | C{2.0cm} |}
\multicolumn{1}{c}{\includegraphics[height=0.08\linewidth, width=0.11\linewidth]{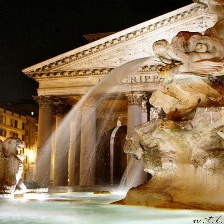}} &
\multicolumn{1}{c}{\includegraphics[height=0.08\linewidth, width=0.11\linewidth]{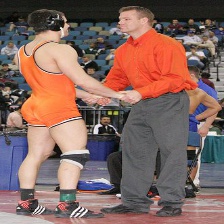}} &
\multicolumn{1}{c}{\includegraphics[height=0.08\linewidth, width=0.11\linewidth]{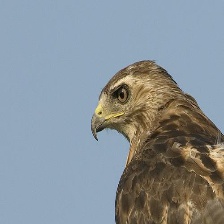}} &
\multicolumn{1}{c}{\includegraphics[height=0.08\linewidth, width=0.11\linewidth]{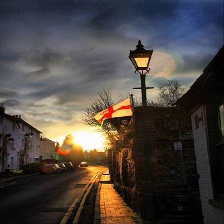}} &
\multicolumn{1}{c}{\includegraphics[height=0.08\linewidth, width=0.11\linewidth]{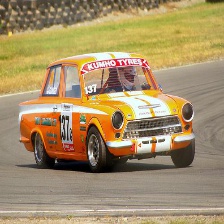}} & \multicolumn{1}{c}{\includegraphics[height=0.08\linewidth, width=0.11\linewidth]{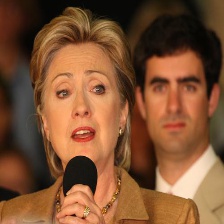}} & \multicolumn{1}{c}{\includegraphics[height=0.08\linewidth, width=0.11\linewidth]{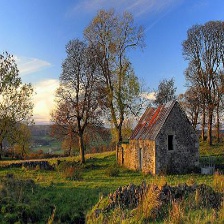}} & \multicolumn{1}{c}{\includegraphics[height=0.08\linewidth, width=0.11\linewidth]{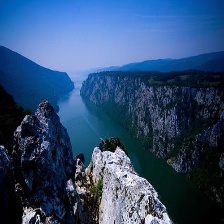}} \\ \hline

{\color{black} water, italy, rome, fountain, roma, italians}  &
{\color{black} man, men, sports, male, sport, university, college, athlete}  & 
{\color{black} bravo, birds, morning, adult, hawk}  & 
{\color{black} sky, street, sun, road, morning,  breathtaking, flag, houses, lamp} & 
{\color{black} car, cars, racing} &
{\color{black} politics, party, election} &
{\color{black} bravo, trees, building, abandoned, wall, stone, farm, fields} &
{\color{black} nature, water, green, landscape, river, park, national, canyon} \\ \hline


{\color{OliveGreen} fountain, italy, water, rome, roma}&
{\color{OliveGreen} college, male, sport, university, athlete}  &
{\color{OliveGreen}  birds, }{\color{red} philadelphia}, {\color{obblue} bird, nature} {\color{OliveGreen} hawk }& 
{\color{obblue} sunset, }{\color{OliveGreen} sun, road, breathtaking, street}&
{\color{OliveGreen} cars, racing, car, }{\color{obblue} classic, auto} & 
{\color{OliveGreen} politics, election, party, }{\color{obblue} democrat}, {\color{red} race} & 
{\color{obblue}cottage, village, clouds, } {\color{OliveGreen} scotland, trees} &
{\color{OliveGreen} water, lake, }{\color{obblue}trees}, {\color{obblue} sun, }{\color{OliveGreen} canoe} \\ \hline
\end{tabular}
\caption{\small{Some example images from the NUS-WIDE dataset with the ground-truth tags in the first row and the top-5 recommended tags in the second row. The tags in green match the ground truth, tags in blue are relevant but not present in the ground truth, and the red ones are wrong predictions.}}
\label{table_sample_results_nus1}
\end{table*}

\paragraph{\textbf{Discriminator}}
A well trained multi-label image classifier can be used to predict tags corresponding to all the visual objects present in an image. This prediction can be fairly accurate and comprehensive in terms of what appears in the image. However, this prediction will be different from how humans will tag the same image. Humans tag their images differently and choose a certain set of tags instead of listing all of the objects present in the image. The choice of tags depends on tagging behavior as well as their personal interests. The task of the discriminator is to differentiate between tags predicted by the network and the actual user-generated tags. The discriminator is trained to maximize the score $\mathcal{D}(T_{gt})$ for actual user tags $T_{gt}$ and minimize the score $\mathcal{D}(\mathcal{G}(\boldsymbol{e}))$ for the generated tags $\mathcal{G}(\boldsymbol{e})$. This can be achieved by optimizing the following loss,
\begin{equation}
\begin{aligned}
L_d &= \max_{\mathcal{D}}\Big(E_{x \sim \boldsymbol{T_{gt}}}[log \mathcal{D}(x)] + \\ 
                    & E_{z \sim \mathcal{F}\boldsymbol(I)}[log(1 - \mathcal{D}(\mathcal{G}(z)))]\Big),
\end{aligned}
\label{equ:D_objective}
\end{equation}
where $L_d$ is the discriminator loss and $\boldsymbol{T_{gt}}$ is the set of actual user-generated tags. The discriminator network consists of a series of fully connected layers (1024, 256, 64, 16) with $relu$ activations which predicts a single value at the end using $sigmoid$ activation.

The ground truth labels associated with an image will either be 0 or 1 based on the absence or presence of any tag. The predicted tags $\mathcal{G}(\boldsymbol{e})$ on the hand will have probabilities corresponding to each label ranging between 0-1. This will make it very easy for the discriminator to distinguish between the network generated and the actual user-generated tags. We address this issue by introducing \textit{jittering} in the actual user-generated tags. The tag labels are modified using the following update rule,
\begin{equation}
\hat{p_i} = (p_i \times \eta) + random(0, \iota),
\label{equ:tag_update}
\end{equation}
where $\hat{p}_i$ is the updated confidence score of a label in user-generated tags for an image $i$, $p_i$ is either 0 or 1 based on the presence or absence of that tag in image $i$, $\eta$ is a threshold which we set to 0.7 in our experiments, $random(0, \iota)$ will generate a random number between 0 and $\iota$, and $\iota$ is a constant which we set to 0.3 in our experiments. This will ensure that the confidence for the presence of a tag will range between (0.7-1.0), and the confidence for the absence of a tag will range from (0.0-0.3). This jittering of user-generated tags was very crucial for training a useful discriminator, which in turn makes the adversarial learning very effective.

\subsection{Implementation and Training}
The proposed framework can be trained end-to-end by joint optimization of the multiple loss functions. We pre-process the image frames similar to \cite{krizhevsky_2012_imagenet} and use a random crop of 224x224 before using the cropped images for training. We implement our code in Keras with Tensorflow backend and use Titan-X GPU for training our network. The full network and the discriminator are trained in an iterative manner, one after the other in each training step. We use ImageNet \cite{krizhevsky_2012_imagenet} pre-trained weights for the ResNet-50 base. We train both the networks with a batch size of 50 using \textit{Adadelta}  optimizer \cite{zeiler2012adadelta} and a learning rate of 1e\textsuperscript{0} until convergence. All the convolution layers (except ResNet) in the network have kernels with size 3x3 and use $relu$ activation. The fully connected layers also use $relu$ activations except for the prediction layers and latent encoding layer, which use $sigmoid$ and $tanh$ activations, respectively.

\begin{table*}[t!]
\centering
\begin{tabular}{ C{2.5cm} | C{1.1cm} | C{.8cm} | C{.8cm} | C{10.3cm}}
\cline{1-5}
\cline{1-5}
\textbf{Dataset} & \textbf{Max} & \textbf{Min} & \textbf{Mean} & \textbf{Tags} \\ \cline{1-5}
NUS-WIDE \cite{chua_2009_nus} & 3348 & 4 & 235 & sky, water, blue, nature, clouds, landscape, explore, sunset, bravo, sea \\ \cline{1-5}
YFCC100M \cite{yfcc100m} & 1017156 & 1881 & 40628 & travel,nature,wedding,vacation, beach,people,art,architecture,summer,party\\ \cline{1-5}
\end{tabular}
\caption{\small{The distribution of the total number of images per tag in NUS-WIDE and YFCC100M dataset we used for training. \textbf{Max}: maximum count of images per user-tag, \textbf{Min}: minimum number of image per user-tag, \textbf{Mean}: mean number of images per user-tag, and \textbf{Tags}: top tags present in the dataset in their frequency order. 
}}
\label{table_data}
\end{table*}


\section{Experiments}
\label{lb_results}

\subsection{Datasets}
\label{lb_dataset}
We perform our experiments on two different datasets, YFCC100M \cite{yfcc100m} and NUS-WIDE \cite{chua_2009_nus}. A summary of the statistics on these two datasets is shown in Table \ref{table_data}.

\paragraph{\textbf{YFCC100M}}
The Yahoo-Flickr Creative Commons 100M (YFCC100M) dataset \cite{yfcc100m} is a large-scale image dataset with around 100 million images. We use the split suggested by \cite{rawat2016contagnet} in our experiments. There are around 28 million images in the training set and around 68K images in the test set. We randomly sample around 2 million images from the $\sim$28 million training set to train our proposed networks. The images in this dataset are annotated using a vocabulary of 1540 user-generated tags. There are 257,275 unique users in the training set, and we use the images (except the images from the test set) to determine the user-preference for image tags.

\paragraph{\textbf{NUS-WIDE}}
We also perform our experiments on the NUS-WIDE dataset \cite{chua_2009_nus}, which is widely used for multi-label classification. This dataset contains 269,648 images from Flickr, which have been annotated for the presence of 81 concepts and 1000 user-generated tags. We use the Flickr API to retrieve the user information for all the images in the dataset. We were able to retrieve around 209,000 images from Flickr, which we used for our experiments. We follow the training and test split suggested in \cite{chua_2009_nus}, which provides us a split of approximately 204,000 training images and around 5000 test images. The dataset has around 50,000 users, and we use annotations for 1000 tags in our experiments. The user history vectors are estimated using the tags present in the images from the training set.
\subsection{Evaluation Metrics}
We use accuracy, precision, and recall to evaluate the proposed methods. We predict top-k labels for the test images and compared the predictions with the ground-truth labels. We compute precision@k (C-P), recall@k (C-R), and accuracy@k at different thresholds, as suggested by \cite{rawat2016contagnet}. We also compute the per-class precision (O-P), recall (O-R), and f1-score (O-F1) measures for different values of k as suggested by \cite{gong2013deep} to compare our method with existing works on NUS-WIDE dataset.

\begin{table}[t!]
\centering
\small
\begin{tabular}{ c | c | c | c }
\hline
\textbf{Methods} & \textbf{Precision@k} & \textbf{Recall@k} & \textbf{Accuracy@k}\\
\hline\hline
Baseline & 0.04 & 0.03 & 0.18 \\ \hline
CNN \cite{krizhevsky_2012_imagenet} & 0.21 & 0.18 & 0.65 \\ \hline
CNN-CTC \cite{he2015convolutional} & 0.24 & {0.20} & {0.72} \\ \hline
PROMPT \cite{shah2016prompt} & 0.23 & 0.19 & 0.69 \\ \hline
ConTagNet \cite{rawat2016contagnet} & 0.25 & 0.20 & 0.71 \\ \hline \hline
\textbf{Proposed}  & \textbf{0.39} & \textbf{0.28} & \textbf{0.84} \\ \hline
\end{tabular}
\caption{\small{A comparison of quantitative evaluation on YFCC100M dataset for k=5.}}
\label{table_results_yfcc1}
\end{table}

\label{lb_results}


\begin{table}[t!]
\small
\centering
\begin{tabular}{m{2.2cm} c | c | c | c | c | c}
\hline
\textbf{Methods} & \textbf{C-P} & \textbf{C-R} & \textbf{C-F1} & \textbf{O-P} & \textbf{O-R} & \textbf{O-F1} \\
\hline
DLSR \cite{lin2014image} & - & - & - & 0.20 & 0.25 & 0.22 \\ \hline
WARP \cite{gong2013deep} & 0.14 & 0.16 & 0.15 & 0.18 & 0.31 & 0.23 \\ \hline
FastTag \cite{chen2013fast} & - & - & - & 0.20 & 0.25 & 0.22 \\ \hline
Softmax \cite{wang_2016_cnn} & 0.14 & 0.18 & 0.16 & 0.17 & 0.29 & 0.22 \\ \hline
CNN-RNN \cite{wang_2016_cnn} & {0.19} & 0.15 & 0.17 & 0.19 & 0.31 & 0.23 \\ \hline
ConTagNet \cite{rawat2016contagnet} & 0.18 & 0.22 & 0.20 & 0.22 & 0.36 & 0.27 \\ \hline
ML-ZSL $\ast$ \cite{lee2018multi} & - & - & - & 0.23 & 0.26 & 0.24 \\ \hline 
ML-ZSL \cite{lee2018multi} & - & - & - & 0.29 & 0.32 & 0.30   \\     \hline \hline
\textbf{Proposed} $\dagger$ & 0.26 & 0.45 & 0.33 & 0.31 & 0.64 & 0.42 \cr   \hline 
\textbf{Proposed} $\ddagger$ & \textbf{0.33} & \textbf{0.52} & \textbf{0.40}  & \textbf{{0.35}} & \textbf{{0.70}} & \textbf{{0.47}} \\ \hline

\end{tabular}
\caption{\small{A comparison of tag prediction results on NUS-WIDE with k = 10. C-P: per-class precision, C-R: per-class recall, C-F1: per-class F1 score, O-P: overall precision, O-R: overall recall, and O-F1: is the overall F1 score. $\ast$ - Generalized ML-ZSL, $\dagger$ - pre-trained with ImageNet, and $\ddagger$ - pre-trained with YFCC100M.}}
\label{table_results_nus1}
\end{table}

\subsection{Results} 
We evaluated our method for different values of k and the precision, recall and accuracy scores on YFCC100M dataset are shown in Table \ref{table_results_yfcc1} and Table \ref{table_results_yfcc2}. We observe that the recall rate goes up (Table \ref{table_results_yfcc2}) as we increase the value of k, which is intuitive as we are able to detect more tags from the ground truth. We achieve a precision of 0.39, recall of 0.28, and an accuracy of 0.84 at k=5. We also evaluated our method on the NUS-WIDE dataset, which is relatively smaller when compared with YFCC100M. We experimented with two different variations where we use either ImageNet pre-trained weights or YFCC100M pre-trained weights. We also computed per-class precision (C-P) and recall (C-R) values to compare with other existing methods. A quantitative evaluation on NUS-WIDE dataset is shown in Table \ref{table_results_nus1}, \ref{table_results_nus2} and \ref{table_results_nus3}. We observe our method has a good recall, which indicates that the network is able to predict the user tags well. We also observe a significant improvement in scores when pre-trained weights from YFCC100M are utilized for training. 

We have shown some example images from YFCC100M as well as NUS-WIDE dataset along with the ground-truth and recommended tags in Table \ref{table_sample_results_yfcc1} and \ref{table_sample_results_nus1}. We can observe that most of the time, the recommended tags are very close to the ground-truth with some cases where they do not match but are relevant to the input image. These new tags (marked in blue) can have two different sources, 1) they were used by the same user in other similar images or 2) they were used by other users with similar tagging behavior. In both scenarios, the recommendation will be useful if the predicted tags are meaningful. We also observe some failure cases which are marked in red. For some of the cases, it is hard to judge whether it is appropriate or not, such as `\textit{philadelphia}' in the third column and `\textit{race}' in the sixth column of Table \ref{table_sample_results_nus1}. However, for other cases the network fails due to confusion in visual appearance where a `\textit{woman}' is predicted as a `\textit{man}' and the `\textit{night street}' is confused with `\textit{cinema}'.

\begin{table}[t!]
\centering
\small
\begin{tabular}{c|ccc}
\hline
\textbf{Methods} & \textbf{Prec@3} & \textbf{Prec@5} & \textbf{Prec@10} \\
\hline
FastTag \cite{chen2013fast} & 0.21 & 0.17 & 0.13 \\
\hline
ConTagNet \cite{rawat2016contagnet} & 0.32 & 0.27 & 0.20 \\
\hline
CLARE \cite{wang2017clare} & 0.21 & 0.16 & 0.14 \\
\hline 
ResNet-50+Preference & 0.20 & 0.17  & 0.13 \\ 
\hline  \hline
\textbf{Proposed} (ImageNet) & 0.60 & 0.47 & 0.31 \\ 
\hline
\textbf{Proposed} (YFCC100M) & \textbf{0.66} &\textbf{ 0.52} & \textbf{0.35} \\
\hline
\end{tabular}
\caption{\small{A comparison of the average precision score (Prec@k) with existing methods at different threshold on NUS-WIDE dataset.}}
\label{table_results_nus2}
\end{table}

\begin{table*}[t!]
\centering
\small
\begin{tabular}{ c | c | c | c | c | c | c | c | c | c | c | c | c | c | c  }
\hline
\multicolumn{6}{c|}{\textbf{Method variation}} & \multicolumn{3}{c|}{\textbf{Precision@k}} & \multicolumn{3}{c|}{\textbf{Recall@k}} & \multicolumn{3}{c}{\textbf{Accuracy@k}} \\
\cline{0-14}
& UP & Adv-I & Adv-P & Joint & Cold Start & k=3 & k=5 & k=10 & k=3 & k=5 & k=10 & k=3 & k=5 & k=10 \\
\hline\hline
A1 & \xmark & \xmark & \xmark & \xmark & \xmark & 0.11  & 0.10  & 0.08  & 0.05 & 0.07 & 0.12 & 0.28 & 0.37  & 0.52  \\ \hline
A2 & \cmark & \xmark & \xmark & \cmark & \xmark & 0.18 & 0.15 & 0.11 & 0.07 & 0.10 & 0.14  & 0.40  & 0.46  & 0.56 \\ \hline
A3 & \cmark & \cmark & \xmark & \xmark & \xmark & 0.34 & 0.29 & 0.22 & 0.15 & 0.21 & 0.30  & 0.63 & 0.72 & 0.82 \\ \hline
A4 & \cmark & \cmark & \xmark & \cmark & \cmark &  0.31 & 0.26 & 0.19 & 0.14 & 0.19 & 0.28 & 0.60  & 0.69 & 0.79  \\ \hline
A5 & \cmark & \xmark & \cmark & \cmark  & \xmark &  0.43 & 0.36 & 0.26 & 0.19 & 0.26 & 0.37 & 0.75 & 0.81 & 0.88  \\ 
\hline \hline
A6 & \cmark & \cmark & \xmark & \cmark  & \xmark  & \textbf{0.46} & \textbf{0.39} &\textbf{ 0.28} & \textbf{0.30} &\textbf{ 0.28} & \textbf{0.39} & \textbf{0.77} & \textbf{0.84 }& \textbf{0.90} \\
\hline
\end{tabular}
\caption{\small{Ablation study showing a comparison of mean precision, recall and accuracy on YFCC100M 1540 tags. UP: user preference, Adv-I: adversarial loss applied independently on generalized tags, Adv-P: adversarial loss applied on personalized tags, Joint: joint training of user preference and tag prediction, and Cold start: ignoring user preference knowledge during inference.}}
\label{table_results_yfcc2}
\end{table*}

\begin{table*}[t!]
\centering
\small
\begin{tabular}{ c | c | c | c | c | c | c | c | c | c | c | c | c | c | c }
\hline
\multicolumn{6}{c|}{\textbf{Method variation}} & \multicolumn{3}{c|}{\textbf{Precision@k}} & \multicolumn{3}{c|}{\textbf{Recall@k}} & \multicolumn{3}{c}{\textbf{Accuracy@k}} \\
\cline{0-14}
& UP & Adv-I & Adv-P & Joint & Cold Start & k=3 & k=5 & k=10 & k=3 & k=5 & k=10 & k=3 & k=5 & k=10 \\
\hline\hline
A1 & \xmark & \xmark & \xmark & \xmark & \xmark & 0.09 & 0.08 & 0.07 & 0.03 & 0.05 & 0.10 & 0.22 & 0.28 & 0.40 \\ \hline
A2 & \cmark & \xmark & \xmark & \cmark & \xmark & 0.20 & 0.17 & 0.13 & 0.11 & 0.15 & 0.21 & 0.41 & 0.51 & 0.62 \\ \hline

A3 & \cmark & \cmark & \xmark & \xmark & \xmark & 0.17  &  0.14 & 0.11 & 0.09  & 0.13  & 0.19  & 0.39  & 0.47 & 0.58 \\
\hline
A4 & \cmark & \cmark & \xmark & \cmark & \cmark & 0.18  & 0.14 & 0.10 & 0.11  & 0.14  & 0.20  & 0.42  & 0.50 & 0.60 \\ \hline 
A5 & \cmark & \xmark & \cmark & \cmark  & \xmark &  0.46 & 0.39 & 0.25 & 0.33 & 0.40 & 0.50 & 0.82 & 0.86 & 0.91  \\ 
\hline \hline
A6 & \cmark & \cmark & \xmark & \cmark  & \xmark & \textbf{0.66} & \textbf{0.52} & \textbf{0.35} & \textbf{0.48} & \textbf{0.58} & \textbf{0.70} & \textbf{0.95} &\textbf{ 0.97} & \textbf{0.98} \\ \hline
\end{tabular}
\caption{\small{Ablation study showing a comparison of mean precision, recall and accuracy on NUS-WIDE 1000 tags. UP: user preference, Adv-I: adversarial loss applied independently on generalized tags, Adv-P: adversarial loss applied on personalized tags, Joint: joint training of user preference and tag prediction, and Cold start: ignoring user preference knowledge during inference.}}
\label{table_results_nus3}
\end{table*}

\begin{table*}[t!]
\centering
\small
\begin{tabular}{ c| c | c |c | c | c | c | c | c | c | c | c }
\hline
\multicolumn{3}{c|}{\textbf{Method variation}} & \multicolumn{3}{c|}{\textbf{Precision@k}} & \multicolumn{3}{c|}{\textbf{Recall@k}} & \multicolumn{3}{c}{\textbf{Accuracy@k}} \\
\cline{0-11} 
Skip Connection & MSE & Huber & k=3 & k=5 & k=10 & k=3 & k=5 & k=10 & k=3 & k=5 & k=10\\
\hline\hline

\xmark & \cmark & \xmark & 0.49 & 0.39 & 0.26 & 0.33 & 0.42 & 0.51  & 0.83 & 0.86 & 0.91 \\ \hline

\cmark & \cmark & \xmark & 0.53 & 0.43 & 0.27 & 0.35 & 0.45 & 0.55  & 0.86 & 0.90 & 0.93 \\ \hline

\xmark & \xmark & \cmark & 0.54 & 0.43 & 0.29 & 0.36 & 0.48 & 0.58  & 0.87 & 0.92 & 0.94 \\ \hline 

\hline \hline
\textbf \cmark  & \xmark & \cmark &  \textbf{0.66} & \textbf{0.52} & \textbf{0.35} & \textbf{0.48} & \textbf{0.58} & \textbf{0.70} & \textbf{0.95} &\textbf{ 0.97} & \textbf{0.98} \\ \hline
\end{tabular}
\caption{\small{Ablation study showing the effect of skip connections and Huber loss in user preference learning.}}
\label{table_results_nus4}
\end{table*}

\subsection{Comparison}
\label{lb_comp}
We compare our proposed method to some baselines and other existing methods on multi-label image classification. We have a baseline (Baseline in Table \ref{table_results_yfcc1}), where we use the most utilized tags for a recommendation. Apart from this, we use the ResNet-50 model with only visual features as a baseline, which is trained using cross-entropy loss on the ground-truth tags. 

We have quantitatively compared our proposed method on the YFCC100M dataset with some of the existing works. YFCC100M is relatively a new dataset which is much bigger and very challenging due to noisy labels and the presence of a large number of users. We compare our method with \cite{krizhevsky_2012_imagenet} and \cite{he2015convolutional} with a modified loss function for multi-label classification. We also compared with recent works on context-aware tag recommendation \cite{shah2016prompt, rawat2016contagnet}, where the authors exploit the presence of meta-data along with images. We observe in Table \ref{table_results_yfcc1} that the proposed method outperforms these works and improves the precision, recall, and accuracy scores by a significant margin.

We also compared our method with existing works on the NUS-WIDE dataset. NUS-WIDE dataset is relatively small but very popular for multi-label classification. However, most of the works mainly focus on annotations with 81 visual concepts, which are lab curated \cite{liu2018multi}. We focus on a much bigger vocabulary (1K tags), which consists of user-generated image tags. We compare our method with DLSR \cite{lin2014image}, WARP \cite{gong2013deep}, CNN-RNN \cite{wang_2016_cnn}, Fast Tagging \cite{chen2013fast}, ConTagNet \cite{rawat2016contagnet}, CLARE \cite{wang2017clare}, and ML-ZSL \cite{lee2018multi}. The comparison results are shown in Table \ref{table_results_nus1} and \ref{table_results_nus2}. Table \ref{table_results_nus1} compares the average precision, recall, and F1 score for the top-10 predicted tags for per-class and overall predictions. We observe that the proposed method outperforms existing works by a significant margin, mainly on recall measure. The same model fine-tuned on the YFCC100M dataset provides an additional boost to the performance. Table \ref{table_results_nus2} shows the precision comparison results for different values of k. Here also, we can observe that our proposed method performs better than the existing methods in terms of precision when tested for different values of k.

The existing approaches utilize tag co-occurrence \cite{wang_2016_cnn}, tag-label relation \cite{wang2017clare} and structural knowledge graph \cite{lee2018multi} to improve the multi-label classification. Personalization plays a big role in a recommendation, which is not considered in these works, and therefore utilizing user-preference helps our approach in predicting better recommendations. The method proposed in \cite{rawat2016contagnet} makes use of contextual information of the captured image, which provides crucial clues useful for tag prediction. However, it also ignores user preference for the recommendation. The work in \cite{shah2016prompt} does utilize user-preference, but the proposed user-preference estimation is independent of visual feature learning. They proposed to create user groups based on tag co-occurrences and utilize the groupings as a user preference. The independent inference of user preference limits the performance of this approach. Our approach learns the user preference jointly with visual encodings leading to a better recommendation. Apart from this, the novel adversarial learning also plays a key role in significant improvement.

\begin{table}[t!]
\centering
\scriptsize
\setlength\tabcolsep{1pt} 
\begin{tabular}{ | C{2.0cm} |  C{2.0cm} | C{2.0cm} | C{2.0cm} | C{2.0cm} | C{2.0cm} |C{2.0cm} |C{2.0cm} |}
\multicolumn{1}{c}{\includegraphics[height=0.16\linewidth, width=0.24\linewidth]{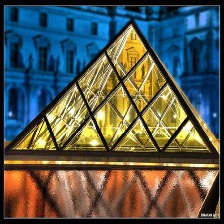}} & \multicolumn{1}{c}{\includegraphics[height=0.16\linewidth, width=0.24\linewidth]{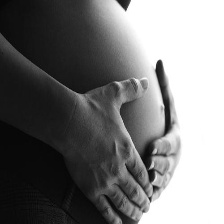}} &
\multicolumn{1}{c}{\includegraphics[height=0.16\linewidth, width=0.24\linewidth]{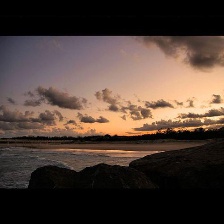}} & \multicolumn{1}{c}{\includegraphics[height=0.16\linewidth, width=0.24\linewidth]{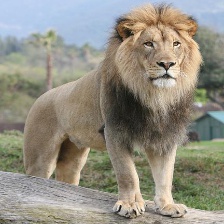}} \\ \hline
{\color{black} architecture, building, paris, golden, pyramid, photographer, france}  & 
{\color{black} people, woman, baby, new, hands, mother}  & 
{\color{black} nature, water, clouds, sunset, beach, ocean, twilight, evening} & 
{\color{black} california, animal, park, wild, cats, south, lion, african}     \\  \hline

{\color{OliveGreen}  france, } {\color{obblue}blue} {\color{red} angle, } {\color{OliveGreen} pyramid, } {\color{obblue}explore}& 
{\color{OliveGreen} baby, } {\color{red} wind,} {\color{obblue}hand}, {\color{red} angel, dress} &
{\color{red} copyright, storm, white, berlin, blue} &
{\color{red} alaska, horse, street, bravo, natural}  \\ \hline

{\color{red}  sky, clouds,} {\color{obblue} reflection}, {\color{red}lamp, england}& 
{\color{red} religion, sports, picture, pink, france} &
{\color{OliveGreen} nature, } {\color{red} white, snow, trees, mist} &
{\color{OliveGreen} animal, } {\color{red}silver,} {\color{obblue}animals,} {\color{red}candid, model}  \\ \hline

{\color{OliveGreen}  pyramid, } {\color{red}cathedral,} {\color{obblue} glass}, {\color{red} sunset, church}& 
{\color{OliveGreen} hands, }{\color{obblue} hand, } {\color{OliveGreen} baby, } {\color{obblue} love,} {\color{red} hospital} &
{\color{OliveGreen} sunset, beach, } {\color{red}sunrise, } {\color{red}lighthouse, dawn} &
{\color{OliveGreen} lion, } {\color{obblue} zoo,} {\color{red}castle, family, wow}  \\ \hline

{\color{OliveGreen}  france, paris, pyramid, architecture, } {\color{obblue} museum}& 
{\color{OliveGreen} baby, hands, } {\color{obblue}love, child, hand} &
{\color{OliveGreen} clouds, ocean, water, beach, rocks} &
{\color{OliveGreen} lion, park, african, animal, wild} 
\\ \hline
\end{tabular}
\caption{\small{A comparison of various models based on the predicted tags for images from NUS-WIDE dataset. The tags shown are top-5 predictions (second row to bottom: ResNet-50, User-preference + Joint, User-preference + Adv-I + Joint + Cold start, User preference + Adv-I + Joint). The tags in green match the ground truth, tags in blue are relevant but not present in the ground truth, and the red ones are wrong predictions.}}
\label{table_sample_results_nus2}
\end{table}

\subsection{Ablation Study}
We performed some ablation experiments to study the effect of various components in our model. The ablation study was performed on both YFCC100M and NUS-WIDE dataset. We use the ResNet-50 model (A1) trained using cross-entropy loss using only visual features as our baseline. The effect of user-preference is shown in A1, where the model integrates user-preference along with visual features (A1 vs A2). The full model, which utilizes adversarial loss along with the integration of user-preference (A6), shows the effect of both adversarial loss and user-preference (A1 vs. A2 vs. A6). To study the impact of joint learning (A3 vs. A6), we trained the full model without joint training (A3). The effect of adversarial learning on personalized tags is also compared with adversarial loss on general tags (A5 vs. A6). We also tested our model for a cold-start scenario where we do not have user tagging behavior for a recommendation (A4 vs. A6). The quantitative evaluation of all the ablation studies is shown in Table \ref{table_results_yfcc2} and \ref{table_results_nus3}. Table \ref{table_sample_results_nus2} and Table \ref{table_sample_results_yfcc4} shows some example images along with the predicted tags for different network configurations. We will discuss and analyze these ablations in the following subsections.

\begin{figure}[t!]
\centering\includegraphics[width=0.70\linewidth]{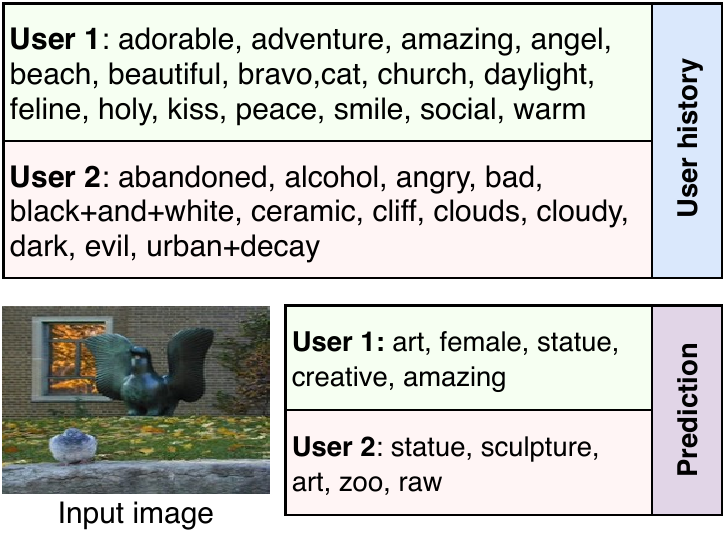}
\caption{\small{Effect of personalization. Here we can observe how different set of tags are recommended to two different users matching their tagging history.}}
\label{fig_pers}
\end{figure}

\paragraph{\textbf{Personalization}}
The integration of user-preference (UP) improve the network's performance to some extent. However, the improvement is significant when we have an additional adversarial loss with joint training. We explore this further and analyze the network prediction to study the effect of personalization. We take two users from NUS-WIDE dataset with different tagging history and analyze the personalized recommendation for the same input image. The input image, along with the predicted tags and user history, is shown in Figure \ref{fig_pers}. We observe that \textit{user1} has a positive and more expressive tagging behavior (\textit{adorable, amazing, angel, etc.}), and \textit{user2} has used mainly used negative and passive image tags (\textit{abandoned, angry, bad, etc.}). We find that the network predicts a different set of tags for these users, which match their tagging behavior to some extent.

\begin{table}[t!]
\centering
\scriptsize
\setlength\tabcolsep{1pt} 
\begin{tabular}{ | C{2.0cm} |  C{2.0cm} | C{2.0cm} | C{2.0cm} | C{2.0cm} | C{2.0cm} |C{2.0cm} |C{2.0cm} |}
\multicolumn{1}{c}{\includegraphics[height=0.16\linewidth, width=0.24\linewidth]{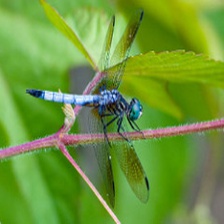}} & \multicolumn{1}{c}{\includegraphics[height=0.16\linewidth, width=0.24\linewidth]{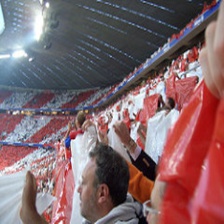}} &
\multicolumn{1}{c}{\includegraphics[height=0.16\linewidth, width=0.24\linewidth]{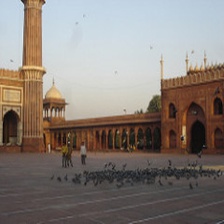}} & \multicolumn{1}{c}{\includegraphics[height=0.16\linewidth, width=0.24\linewidth]{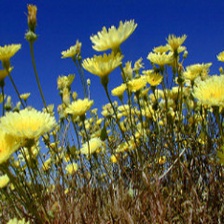}} \\ \hline
{\color{black} aquarium, dragonfly, insect, male, nature, stream, wildlife}  & 
{\color{black} arena, football, rot, soccer, stadium}  & 
{\color{black} holiday, mosque, travel, trip, vacation} & 
{\color{black} dandelion, desert, flower, nature, spring, yellow}     \\  \hline

{\color{OliveGreen}nature, \color{red}flower, \color{obblue}macro, \color{OliveGreen}insect, \color{red}garden}& 
{\color{obblue}people, \color{red}car, travel, parade, food} &
{\color{OliveGreen}travel, \color{red}water, beach, winter, snow} &
{\color{red} travel, \color{OliveGreen}nature, \color{red}architecture, landscape, summer}  \\ \hline

{\color{OliveGreen}nature, wildlife,\color{red} bay, red, vacation}& 
{\color{red} wales, \color{obblue} red, \color{red}gig, bird, diving} &
{\color{OliveGreen}travel, \color{obblue}sky, \color{red}clouds, tree, sunrise} &
{\color{red}landscape, panorama, \color{obblue}blue, \color{red}night, \color{obblue}green}  \\ \hline

{\color{OliveGreen}dragonfly, \color{obblue}macro, \color{OliveGreen}insect, \color{red}fly, \color{obblue}bug}& 
{\color{OliveGreen}football, soccer, stadium, \color{obblue}crowd, \color{red}baseball} &
{\color{red}sunset, \color{obblue}architecture, \color{OliveGreen}mosque, \color{red}statue, \color{obblue}building} &
{\color{OliveGreen}yellow, flower, spring, nature, \color{obblue}blue}  \\ \hline

{\color{OliveGreen}nature, dragonfly, wildlife, \color{OliveGreen}insect, \color{obblue}macro}& 
{\color{OliveGreen}football, soccer, stadium, \color{red}festival, \color{obblue}crowd} &
{\color{OliveGreen}travel, trip, vacation, holiday, \color{obblue}architecture} &
{\color{OliveGreen}flower, spring, yellow, \color{obblue}blossom, \color{OliveGreen}nature} 
\\ \hline
\end{tabular}
\caption{\small{A comparison of various models based on the predicted tags for images from YFCC100M dataset. The tags shown are top-5 predictions (second row to bottom: ResNet-50, User-preference + Joint, User-preference + Adv-I + Joint + Cold start, User preference + Adv-I + Joint). The last three models use joint training. The tags in green match the ground truth, tags in blue are relevant but not present in the ground truth, and the red ones are wrong predictions.}}
\label{table_sample_results_yfcc4}
\end{table}

We also perform some ablation experiments to study the effect of skip connections and Huber loss in the user-preference learning. We use the full proposed network for these ablations with changes in the user preference network. All the experiments were performed on the NUS-WIDE dataset with 1000 tags. A comparative evaluation is shown in Table \ref{table_results_nus4}. We can observe that adding skip connections and the use of Huber loss, both help in improving network performance.  

\paragraph{\textbf{Effectiveness of Adversarial Loss}}
We observe that the integration of user-preference with visual encodings does help in improving the performance of the model. However, training with an adversarial loss (Adv-I) provides a significant boost to the network's performance. We observe a $\sim$20-40\% improvement in almost all the evaluation metrics after integrating the adversarial loss in the network for the YFCC100M dataset. We also observe that the improvement on NUS-WIDE was almost similar to the gain with user preference. The main reason for this could be the presence of a large number of users in YFCC100M, which provides sufficient variation for adversarial learning.

\begin{table*}[t!]
\centering
\scriptsize
\setlength\tabcolsep{4pt} 
\begin{tabular}{  C{1.2cm} |  C{2.5cm} | C{2.5cm} | C{2.5cm} | C{2.5cm} | C{2.5cm} |C{2.0cm} |}
\hline


User history tags & None  & 
beach, friends, hdr, raw, sea, white &
cat, dog, family, festival, food, girl, graffiti, macro, music, wedding & 
architecture, art, clouds, flower, garden, holiday, live, portrait, snow & 
blue, car, city, fun, green, landscape, life, nature, park, snow, summer, tree, winter \\ \hline

Image & Cold Start (0\%) & 25\% & 50\% & 75\% &  100\% \\ 
\cline{2-6}

 {\includegraphics[height=0.75\linewidth, width=1\linewidth]{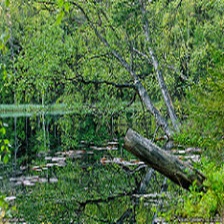}} & 
 forest, green, nature, tree, water & 
 water, vacation, green, summer, nature & 
 vacation, nature, beach, water, family & 
 nature, vacation, raw, water, love & 
 nature, summer, water, green, light \\ 
 \cline{2-6}
 {\includegraphics[height=0.75\linewidth, width=1\linewidth]{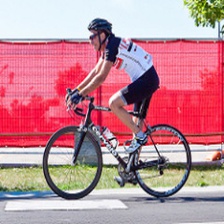}} &  
 speed, cycling, bike, helmet, bicycle & 
 bike, speed, fun, fast, race & 
 bike, fun, speed, beach, vacation & 
 fun, bike, speed, beach, race & 
 bicycle, bike, fun, event, person  \\ 
 \cline{2-6}
\end{tabular}
\caption{\small{Effect of variation in user history on the recommended tags. As the user tag history evolves, we can observe the variation in the recommended tags, which are predicted using the proposed model on samples from YFCC100M dataset. The first row shows tags which are added at each step to the user history (top frequent). The recommendations (row 3-4 and column 2-6) are the top 5 predictions ordered based on the confidence score.}}
\label{table_dynamic_history}
\end{table*}

We also perform an ablation where the adversarial loss was computed over personalized tags (Adv-P) instead of independent generalized tags. We observe that the improvement is more significant when the adversarial learning is performed independently from user preference learning (last row vs. second last row in Table \ref{table_results_yfcc2} and Table \ref{table_results_nus3}). There are two main reasons for this. 1) The personalized tags are conditioned on user preference, which already helps the predictions to be close to the ground-truth tags (please note that the ground-truth tags are personalized). On the other hand, the generalized tags have no notion of personalization, and the predicted tags will not be that close to the ground-truth. The adversarial loss has a better scope of learning when the discriminator can easily distinguish between the real and fake predictions as compared to the case when real and fake predictions are already very close. 2) The user preference and visual encoding are learned in joint training, which is found to be effective. If adversarial learning is performed on the personalized tags, the adversarial loss will also affect the learning of user preference. However, the adversarial loss should not update the user preference as the real/fake tags are not paired based on users, and they could be from entirely different users. The adversarial loss only focuses on whether the distribution of the predicted tags is similar to personalized tags irrespective of the user.

\paragraph{\textbf{Joint Training}}
We perform our experiment on both NUS-WIDE and YFCC100M dataset to study the effect of joint training on the performance of the proposed network. The results are shown in Table \ref{table_results_yfcc2} and Table \ref{table_results_nus3} (UP + Adv-I and UP + Adv-I + Joint). We observe that joint training improves the performance significantly. This indicates that the joint training is important for learning a meaningful visual as well as user-preference encoding, which in turn is effective for tag prediction.

\paragraph{\textbf{Cold Start Users}}
We further analyzed the performance of the proposed method for cold start scenarios where we do not have any user history. We set the user history vector to all zeros in this case and evaluated the network prediction. The evaluation results are shown in Table \ref{table_results_yfcc2} and \ref{table_results_nus3} (UP + Adv-I, Cold Start). We observe that the network performs significantly better than the baselines and existing methods on the YFCC100M dataset, indicating the usefulness of the proposed method. We also observe that the network does not perform so well on the NUS-WIDE dataset in comparison with YFCC100M. This variation could be due to the difference in the size of these datasets in terms of the number of images and number of users.

\paragraph{\textbf{Dynamic User History}}
User tagging history evolves over time, and therefore, the recommendation should also change with this variation. The proposed approach utilizes tagging history for the recommendation; therefore, it can also be used for dynamic user history without any modifications. We performed an experiment where we use an evolving user history for generating the recommendation. We start with a cold-start scenario with no user history and keep adding tags, until we have all the user tags, as we generate recommendations. The results from this experiment are shown in Table \ref{table_dynamic_history}. We can observe that as the user history evolves, the recommendation also changes. This is an interesting aspect of recommendation which requires further in-depth analysis and can be a promising research direction.

\section{Conclusion and Future Work}
\label{lb_conclusion}
In this work, we propose a unified deep network for preference-aware tag recommendation, which can be trained end-to-end on a large-scale dataset. We use an encoder-decoder network equipped with skip connections, which enables efficient and \textit{unsupervised} user-preference learning. A joint training of user-preference and visual encoding allows the network to efficiently integrate the visual preference with tagging behavior for a better recommendation. We also propose the use of \textit{adversarial} learning to further enhance the quality of tag prediction. The adversarial loss enforces the network to learn features which can be used to predict tags with a distribution which resembles human-generated tags. We perform extensive experiments on two different datasets, YFCC100M and NUS-WIDE. The proposed methods achieve significant improvement on the large-scale YFCC100M dataset and also outperform existing methods on the NUS-WIDE dataset. In future work, we plan to explore the proposed network for the task of preference-aware image captioning.

\bibliographystyle{IEEEtran}
\bibliography{IEEEabrv,sigproc}

\end{document}